\documentclass{article}
\usepackage{spconf,amsmath,amssymb,bm,graphicx,float,caption,hhline,fancyhdr}
\usepackage[hyphens]{url}
\usepackage{xcolor}
\usepackage{soul}
\usepackage[colorlinks,citecolor=green,urlcolor=blue,bookmarks=false,hypertexnames=true,linkcolor=red]{hyperref} 
\floatstyle{plaintop}
\restylefloat{table}
\usepackage{multirow}
\usepackage{makecell}
\usepackage{array}

\DeclareMathOperator*{\argmax}{arg\,max}

\title{Boosting Image Quality Assessment Performance: Unsupervised \\Score Fusion by Deep Maximum a Posteriori Estimation}

\name{Zhongling Wang$^{\star}$, Raymond Zhou$^{\star}$, Shahrukh Athar$^{\dagger}$, Wenbo Yang$^{\star}$, and Zhou Wang$^{\star}$}
\address{$^{\star}$University of Waterloo, Canada \;\; $^{\dagger}$ McMaster University, Canada}

\begin{document}
\ninept
\maketitle

\begin{abstract}
Over the past decades, numerous Image Quality Assessment (IQA) models have emerged, aiming to predict the perceptual quality of images. However, individual models are often biased toward certain types of image content or distortions, depending on the design principle and process. An intuitive idea is to harness the strengths and mitigate the weaknesses of each IQA model, by fusing the scores of multiple models into a stronger one. Here we make one of the first attempts to seek an optimal solution for the idea and propose a general framework for unsupervised IQA score fusion using deep Maximum a Posteriori (MAP) estimation. The proposed model conducts fine-grained uncertainty estimation at the score level to increase the accuracy and reduce the uncertainty in fused predictions. Comprehensive experiments demonstrate the superiority of the proposed model over individual IQA models and other fusion methods. It also exhibits an interesting capability of rejecting ``bad" models in the fusion process.

\end{abstract}


\section{Introduction}

Image Quality Assessment (IQA) models are designed to predict the perceptual quality of images. Over the past decades, numerous IQA models have been introduced. Their correlation with human evaluations, typically represented by the Mean Opinion Score (MOS), has progressively increased. Many IQA models demonstrate superior performance on average when being evaluated on specific datasets. However, due to their distinct design philosophies and training methodologies, they often capture some particular types of distortions or handle some specific image contents better than others. Consequently, individual IQA models often fail to address all types of images and distortions encountered in real-world scenarios. 
While designing such a strong universal model remains a grand challenge, an intuitive idea to quickly boost IQA performance without developing a new one is to leverage existing IQA models by fusing their scores to attain more reliable predictions, so as to harness the strengths and mitigate the weaknesses in each model. 

Existing fusion approaches can be categorized non-mutually exclusively into empirical, rank fusion methods, and supervised learning-based \cite{Shahrukh_review}. Empirical models \cite{fusion_hfsim, fusion_cisi, fusion_cm, fusion_cm_multi, fusion_gen_ref1, fusion_cqm, fusion_ehis, fusion_bmmf1} fuse a predetermined set of IQA models using a handcrafted formula. This approach significantly constrains its adaptability when introduced with new IQA models. Rank fusion methods operate in the discrete rank domain, where the range of all IQA models is mapped to the same uniform distribution. However, these methods are tied closely to the diversity of the ranking dataset, which can impede generalizability. Supervised learning-based methods \cite{fusion_mmf_tip, fusion_md_svr} are trained under the guidance of the MOS of a single subjective rated dataset. Such fusion methods are essentially refined versions of supervised learning-based IQA models since they share the same ground truth, i.e., MOS of a specific dataset, as the base IQA models.

\begin{figure}[!ht]
\centering
\includegraphics[clip, trim=3cm 5.5cm 13cm 3.8cm, width=0.5\textwidth]{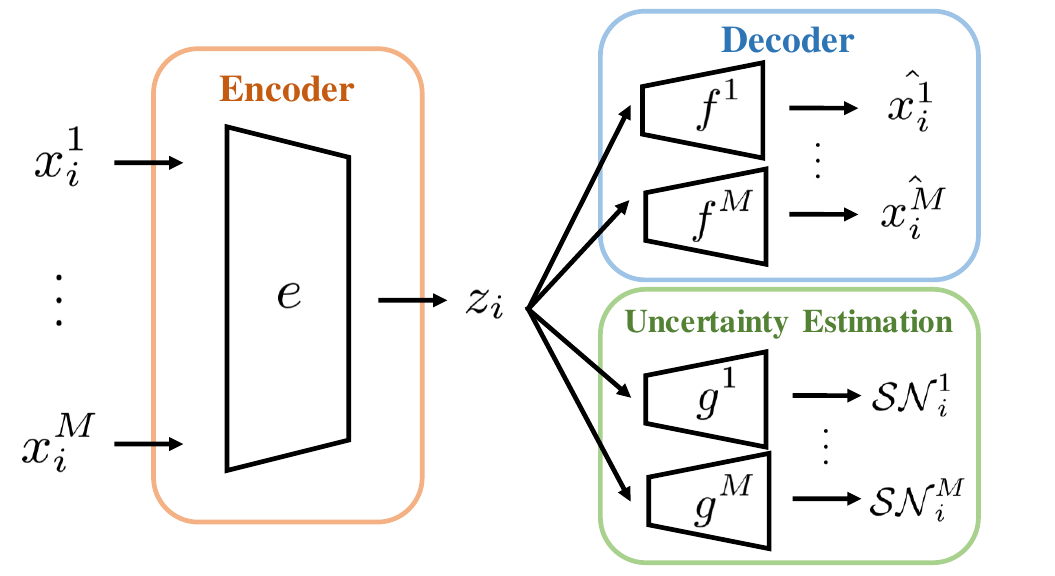}
\caption{The diagram of the proposed framework fusing IQA scores $\mathbf{x_i} = \{x_i^j | j=1, 2,...,M\}$ of image $I_i$. $\hat{x_i^j} = f^j(z_i)$ represents the reconstructed score. $\mathcal{SN}_i^j$ is the abbrev.\ of $\mathcal{SN}(\xi_i^j, \omega_i^j, \alpha_i^j)$,  which is the predicted conditional Skew Normal distribution $p(x_i^j | z_i)$.} 
\label{fig:framework}
\end{figure}

We argue that supervised-learning based fusion approaches, i.e., when MOSs are used as the guidance, in essence, counter the fundamental reasoning behind the score fusion idea in its attempts to improve generalizability, because any MOS is associated with some specific image content and specific distortion types and levels, and thus inevitably leads to biases that we try to avoid. By contrast, without using MOS, or unsupervised score fusion, is inherently a completely different and more challenging problem. 
In this work, we focus on unsupervised score fusion \textit{without} involving MOS. We believe that the key to the problem is to estimate the uncertainty of each IQA score without any ground truth. Existing fusion methods either lack uncertainty estimation or only remain coarse-grained at the model level. To mitigate this problem, we proposed a general framework for unsupervised IQA score fusion using Maximum a Posteriori (MAP) estimation. Our framework consists of an encoder, a set of decoders and a set of uncertainty estimation modules. The encoder fuses individual IQA models, which not only expedites inference but also enhances the framework's explainability. The decoders model the relationship between MOS and each IQA model. Uncertainty estimation modules are responsible for fine-grained, score-level uncertainty estimation. Given its unsupervised nature, the proposed model demonstrates superior generalizability to unseen data. The fine-grained uncertainty estimation module predicts the uncertainty of each score generated by individual IQA models. By identifying and fusing the less uncertain portions of each model, we achieve a more accurate and reliable prediction. The proposed method is trained end-to-end to ensure all modules collaborate seamlessly. An overview of the framework is shown in Fig. \ref{fig:framework}. Comprehensive experiments on ten testing sets demonstrate the superiority of the proposed model over other ones. 

The main novelties of our work include: \textbf{(1)} To the best of our knowledge, we propose the first unsupervised learning-based score fusion approach for IQA. \textbf{(2)} We formalize the first observation model of IQA fusion and address the task using MAP estimation. \textbf{(3)} By building a powerful fine-grained uncertainty estimation module, the proposed model increases accuracy and reduces uncertainty in its prediction by harnessing the strengths and mitigating the weaknesses of each model. \textbf{(4)} We show that rank fusion can be easily integrated into our general framework.
\textbf{(5)} The proposed model exhibits capability of rejecting ``bad" models in the fusion process.

\section{Related Works}
The first category is empirical score fusion methods. They include earlier methods in IQA score fusion, including HFSIMc \cite{fusion_hfsim}, CISI \cite{fusion_cisi}, CM \cite{fusion_cm,fusion_cm_multi,fusion_gen_ref1}, CQM \cite{fusion_cqm}, EHIS \cite{fusion_ehis} and BMMF \cite{fusion_bmmf1}, etc. Utilizing a predetermined set of IQA methods, scores are fused through a handcrafted weighted sum or weighted product formula. The weights are determined either via prior knowledge or optimization on subjectively rated datasets. However, such methods pose challenges when integrating new IQA models and often fail to account for the score-dependent uncertainties.


Instead of fusing continuous scores, rank fusion methods first convert scores of individual IQA methods to discrete ranks and then fuse them to attain unified rankings or scores. In score fusion, we need to account for the different ranges of these IQA methods, which intensifies the complexity of this task. A notable advantage of converting scores to rankings is that it nonlinearly transforms arbitrary score distributions into the same uniform distribution. RAS \cite{fusion_bliss} is an empirical rank fusion method that fuses the ranks using a handcrafted Reciprocal Rank Fusion (RRF) \cite{fusion_rrf} formula. It is also the only method that applies rank fusion in the IQA field. Subsequent methods described are designed for general purposes or other tasks and have not found application in the area of IQA. Weighted versions of RRF \cite{fusion_weighted_rrf1, fusion_weighted_rrf2} are introduced where the weight is learned by minimizing model-level uncertainty. Distance-based rank aggregation methods \cite{fusion_em_rank2} define distance measures in the discrete space and fuse the ranks by maximizing the likelihood function. However, existing rank fusion methods either lack uncertainty estimation or only remain coarse-grained at the model level. Moreover, the conversion from score to rank inherently results in information loss, setting an implicit upper bound on the optimal performance achievable by these rank fusion methods. An additional limitation is the accuracy of rankings relies heavily on the diversity of the dataset. Optimization in the discrete space is also more time-consuming compared to those in the continuous space.


\section{Maximum a Posteriori Estimation}

\begin{figure}
\centering
\includegraphics[clip, trim=0.2cm 0.4cm 0.2cm 0.2cm,width=0.45\textwidth]{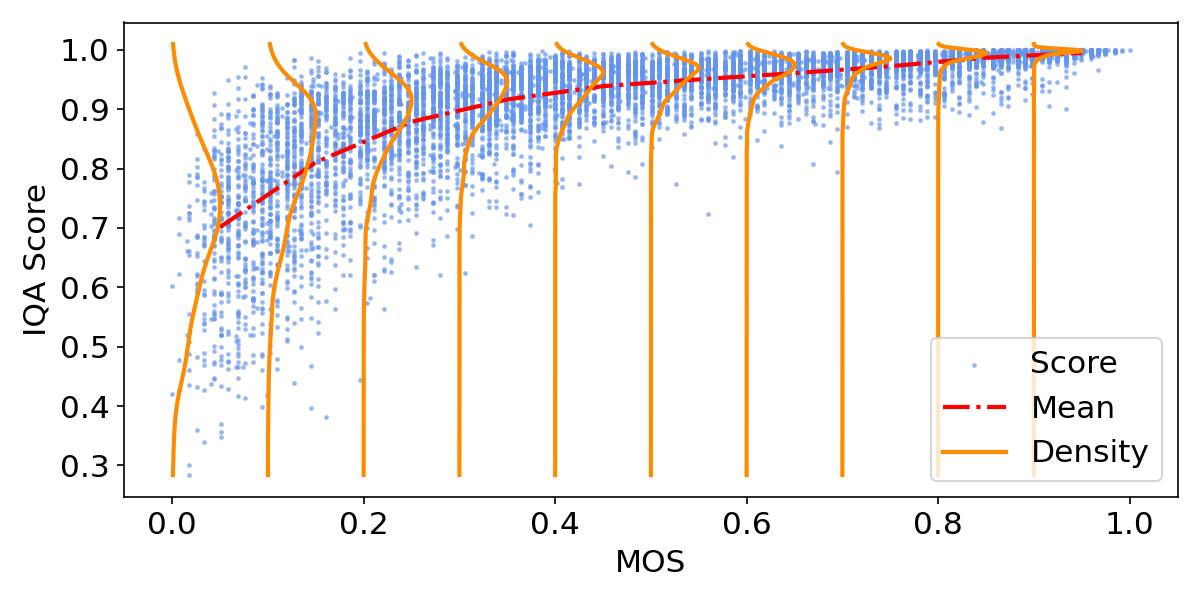}
\caption{The empirical distribution of IWSSIM \cite{fr_iwssim} evaluated on KADID-10K \cite{db_kadid10k}dataset. The dashed red curve is the mean value of the scores (shown as blue dots). The solid orange curves are the conditional distributions, each representing the density of scores given a MOS range. It is easy to find that the conditional distributions are skewed toward the higher score side. Also, the variance of the conditional distribution decreases as the MOS gets higher.}
\label{fig:cond_hist}
\end{figure}

\subsection{Observation Model}

Given $N$ distorted images $\{I^d_i|i=1,2,\dots, N\}$ and their corresponding pristine ones $\{I^r_i|i=1,2,\dots, N\}$, we can evaluate the quality of the distorted images using a Full Reference (FR) IQA metric: $x_i^j = \textit{FR-IQA}_j (I^d_i, I^r_i)$ or a No Reference (NR) one: $x_i^j = \textit{NR-IQA}_j (I^d_i)$. The proposed framework is capable of fusing IQA metrics of all kinds, either FR-IQA or NR-IQA. Suppose that we have scores $\{x_i^j \in \mathbb{R} | i=1,2,\dots, N; j=1,2,\dots,M\}$ generated using $M$ IQA metrics. $x_i^j$ is the score of the $i^{th}$ image evaluated by the $j^{th}$ metric. We adopt the notation where subscripts represent images and superscripts denote models. Our goal is to fuse these scores into final predictions $\{z_i \in \mathbb{R} | i=1,2,\dots, N\}$ that align with MOS better than any individual IQA metric. We assume that the score $x_i^j$ is generated by a function of $z_i$ and two independent noise: a score-dependent noise $n_i^j$ and a model-dependent noise $\hat{n^j}$:

\begin{equation}
\vspace{-0.1cm}
    x_i^j = f^j(z_i) + n_i^j + \hat{n^j}
\label{eq:observation}
\end{equation}

\noindent where$f^j: \mathbb{R} \to \mathbb{R}$ is a deterministic function that simulates the mapping from MOS to score $x_i^j$. $n_i^j$ is a score-dependent noise that follows a skew normal distribution: $n_i^j \sim \mathcal{SN}(\xi_i^j, \omega_i^j, \alpha_i^j)$ where $\xi_i^j$, $\omega_i^j$, $\alpha_i^j$ are the location, scale and shape parameters, respectively. $\omega_i^j$ is determined through a score-level uncertainty estimation function $g^j: \mathbb{R} \to \mathbb{R}$: $\omega_i^j = g^j(z_i)$. For simplicity, we assume $\xi_i^j = 0$ and $\alpha_i^j$ is only model-dependent, therefore we denote it as $\alpha^j$ throughout the paper. $\hat{n^j}$ is a model-dependent noise that follows a zero mean normal distribution: $\hat{n^j} \sim \mathcal{N}(0, {\sigma^j}^2)$. Since $n_i^j$ and $\hat{n^j}$ are independent, we will show later that $n_i^j + \hat{n^j}$ also follows a skew normal distribution. We assume $f^j$ and $g^j$ follow some parametric forms with parameters $\theta_{f^j}$ and $\theta_{g^j}$, respectively.

The motivation behind modeling uncertainty as score-dependent arises from the human visual system's increased uncertainty when evaluating lower-quality images. Similar behavior is also found in many IQA models, showing greater variance in regions of lower MOS. As a result, it is more accurate to model the uncertainty as score-dependent. Furthermore, according to the empirical observation on multiple IQA models, we assume $g_j$ is nondecreasing w.r.t. $z$. This is demonstrated in the variance of the fitted conditional density curves shown in Fig. \ref{fig:cond_hist}. The reason for modeling the conditional distribution $p(x_i^j | z_i)$ as asymmetric is also based on the observation across various IQA models: $p(x_i^j | z_i)$ is usually skewed towards the lower score side. We choose skew normal due to computational feasibility. This characteristic is also illustrated in the fitted conditional density curves in Fig. \ref{fig:cond_hist}. 

\subsection{MAP Formulation and Optimization}
Our method is based on the Maximum a Posteriori (MAP) estimation framework. It's easy to show that the likelihood $p(x_i^j | z_i) = p(n_i^j + \hat{n^j} | z_i)$ is also a skew normal distribution $\mathcal{SN}(0, \tilde{\omega_i^j}, \tilde{\alpha^j})$. The likelihood function can be written as
\vspace{-0.2cm}
\begin{equation}
\begin{split}
    & p(x_i^j | z_i; \tilde{\alpha^j}, \theta_{f^j}, \theta_{g^j}) \\
    = & \frac{\sqrt{2}}{\sqrt{\pi} \tilde{\omega_i^j}} \left(\frac{1}{2} + \frac{\operatorname{erf}{\left(\frac{ \tilde{\alpha^j} \left( n_i^j + \hat{n^j} \right)}{\sqrt{2} \tilde{\omega_i^j}} \right)}}{2}\right) e^{- \frac{(n_i^j + \hat{n^j})^2}{2 \tilde{\omega_i^j}}}
\label{eq:likelihood}
\end{split}
\end{equation}

\noindent where $\tilde{\omega_i^j} = \sqrt{{\omega_i^j}^2 + {\sigma^j}^2}$ and $\tilde{\alpha^j} = \frac{\alpha_i^j \omega_i^j}{\sqrt{ {\omega_i^j}^2 + {\sigma^j}^2 + {\alpha^j}^2 {\sigma^j}^2 }}$ are the new scale and shape parameters, respectively. $\operatorname{erf}$ is the error function defined as $\operatorname{erf}(x) = {\frac {2}{\sqrt {\pi }}}\int _{0}^{x}e^{-t^{2}}\,\mathrm {d} t.$ The MAP objective can be written as 

\begin{align}
&\argmax_{\mathbf{z}, \bm{\alpha}, \bm{\sigma}, \bm{\theta_{f}}, \bm{\theta_{g}}} p(\mathbf{z} | \mathbf{X}; \bm{\alpha}, \bm{\sigma}, \bm{\theta_{f}}, \bm{\theta_{g}}) \nonumber \\
=& \argmax_{\mathbf{z}, \bm{\alpha}, \bm{\sigma}, \bm{\theta_{f}}, \bm{\theta_{g}}} \log p(\mathbf{X} | \mathbf{z}; \bm{\alpha}, \bm{\sigma}, \bm{\theta_{f}}, \bm{\theta_{g}}) + \log p(\mathbf{z}) \label{eq:map}  \\ 
=& \argmax_{\begin{array}{c} z_i, \alpha^j, \sigma^j \\\theta_{f^j}, \theta_{g^j} \end{array} } \sum_j^M \sum_i^N \left( \log p(x_i^j | z_i; \alpha^j, \sigma^j, \theta_{f^j}, \theta_{g^j}) + \log p(z_i) \right) \label{eq:map_factorized}
\end{align}

\vspace{-0.2cm}
\noindent where $\mathbf{X} \in \mathbb{R}^{N \times M}$ represents the IQA scores of the $N$ images calculated by $M$ IQA models. $\mathbf{z} \in \mathbb{R}^N$ is the MOS of the corresponding $N$ images. $\bm{\alpha} \in R^M$, $\bm{\sigma} \in R^M$ are the collections of distributional parameters $\alpha^j, \sigma^j$, respectively. $\bm{\theta_{f}}, \bm{\theta_{g}}$ are the collections of model parameters of $f^j, g^j$, respectively. $p(\mathbf{z})$ is the prior distribution over the predicted MOS $\mathbf{z}$. To make Eq. \ref{eq:map} tractable, we introduce two commonly made assumptions. First, IQA models are conditionally independent, i.e., $p(\mathbf{x_i} | z_i) = \prod_{j=1}^M p(x_i^j | z_i)$ where $\mathbf{x_i} \in \mathbb{R}^M$ are the scores of the same image evaluated by $M$ IQA metrics. Second, the images are independent, that is, $p(\mathbf{X}) = \prod_{i=1}^N p(\mathbf{x}_i)$. Now we can factorize Eq. \ref{eq:map} into a summation of individual loss functions as shown in Eq. \ref{eq:map_factorized}. By substituting Eq. \ref{eq:likelihood} into Eq. \ref{eq:map_factorized}, we get the final objective. The prior distribution $p(\mathbf{z})$ describes our understanding of the dataset. However, given the variability of MOS distributions across datasets, the use of informative priors might compromise generalizability. Without loss of generality, we define $p(\mathbf{z}) = \mathcal{U}(0, 1)$ since most subjective rated datasets advertise themselves as diverse where the MOS is close to uniform in practice \cite{Shahrukh_review}. This prior is used to regularize the range of the predicted $\mathbf{z}$ without compromising generalizability. Finally, we jointly optimize $\mathbf{z}, \bm{\alpha}, \bm{\sigma}, \bm{\theta_{f}}, \bm{\theta_{g}}$ end-to-end.

\begin{table*}[th!]
\scriptsize
\caption{Evaluation results of the proposed model, five other fusion methods and 16 individual IQA models used in the fusion. The average SRCC and PLCC values are computed as weighted sums of individual ones, with weights determined by the number of images in each set. Only SRCC scores are provided for individual testing sets due to the page limit. The top three best-performing models are shown in bold font.}
\vspace{-0.2cm}
\label{table:all}
\centering
\def\tablew{0.76cm}
\begin{tabular}{| >{\centering\arraybackslash}m{1.4cm} | >{\centering\arraybackslash}m{1.5cm} | >{\centering\arraybackslash}m{\tablew} | >{\centering\arraybackslash}m{\tablew} | >{\centering\arraybackslash}m{\tablew} | >{\centering\arraybackslash}m{\tablew} | >{\centering\arraybackslash}m{\tablew} | >{\centering\arraybackslash}m{\tablew} | >{\centering\arraybackslash}m{\tablew} | >{\centering\arraybackslash}m{\tablew} | >{\centering\arraybackslash}m{\tablew} | >{\centering\arraybackslash}m{\tablew} | >{\centering\arraybackslash}m{\tablew} | >{\centering\arraybackslash}m{\tablew} |}
\Xhline{2\arrayrulewidth}
\textbf{Type} & \textbf{Model} & \textbf{LIVEr2} & \textbf{TID2013} & \textbf{CSIQ} & \textbf{VCL @FER} & \textbf{CIDIQ 50} & \textbf{CIDIQ 100} & \textbf{MDID} & \textbf{MDID 2013} & \textbf{LIVE MD} & \textbf{MDIVL} & \textbf{avg SRCC} & \textbf{avg PLCC} \\
\Xhline{2\arrayrulewidth}
\multirow{12}{*}{$\begin{array}{c}\text{Fused} \\ \text{FR-IQA}\end{array}$} 
& VSI\cite{fr_vsi} & 0.9524 & \textbf{0.8965} & 0.9422 & 0.9317 & 0.7213 & 0.8106 & 0.8569 & 0.5700 & 0.8414 & 0.8269 & 0.8631 & 0.8714\\
& FSIMc\cite{fr_fsimc_ref_yiq} & 0.9645 & 0.8510 & 0.9309 & 0.9323 & 0.7608 & 0.8285 & 0.8904 & 0.5806 & 0.8666 & 0.8613 & 0.8628 & 0.8787\\
& FSIM\cite{fr_fsim} & 0.9634 & 0.8015 & 0.9242 & 0.9178 & 0.7438 & 0.8149 & 0.8872 & 0.5817 & 0.8635 & 0.8585 & 0.8628 & 0.8687\\
& IWSSIM\cite{fr_iwssim} & 0.9567 & 0.7779 & 0.9212 & 0.9163 & \textbf{0.8484} & \textbf{0.8564} & 0.8911 & \textbf{0.8551} & \textbf{0.8836} & 0.8588 & 0.8559 & 0.8786\\
& SFF\cite{fr_sff} & 0.9649 & 0.8513 & \textbf{0.9627} & 0.7738 & 0.7834 & 0.7689 & 0.8396 & 0.8005 & 0.8700 & 0.8535 & 0.8527 & 0.8658\\
& DSS\cite{fr_dss} & 0.9616 & 0.7921 & 0.9555 & 0.9272 & 0.7755 & 0.8246 & 0.8658 & \textbf{0.8078} & 0.8714 & \textbf{0.8759} & 0.8520 & 0.8757\\
& QASD\cite{fr_qasd} & 0.9629 & 0.8674 & 0.9530 & 0.9231 & 0.7307 & 0.8079 & 0.7778 & 0.6687 & 0.8766 & 0.8315 & 0.8482 & 0.8638\\
& MCSD\cite{fr_mcsd} & 0.9668 & 0.8089 & \textbf{0.9592} & 0.9224 & 0.7562 & 0.7808 & 0.8451 & 0.8269 & 0.8517 & 0.8370 & 0.8464 & 0.8705\\
& CID MS\cite{fr_cid} & 0.9103 & 0.8314 & 0.8789 & 0.9366 & 0.8350 & 0.8062 & 0.8330 & 0.6168 & 0.8608 & \textbf{0.8778} & 0.8445 & 0.8511\\
& GMSD\cite{fr_gmsd} & 0.9603 & 0.8044 & 0.9570 & 0.9177 & 0.7427 & 0.7675 & 0.8613 & 0.8283 & 0.8448 & 0.8210 & 0.8433 & 0.8672\\
& VIF\cite{fr_vif} & 0.9636 & 0.6769 & 0.9194 & 0.8866 & 0.7203 & 0.6257 & \textbf{0.9306} & \textbf{0.8444} & 0.8823 & 0.8381 & 0.8024 & 0.8388\\
& VIF DWT\cite{fr_dwtvif} & 0.9681 & 0.6439 & 0.9020 & 0.8930 & 0.7224 & 0.5826 & 0.8943 & 0.7553 & 0.8479 & 0.8243 & 0.7768 & 0.8220\\
\hline
\multirow{4}{*}{$\begin{array}{c}\text{Fused} \\ \text{NR-IQA}\end{array}$}
& HOSA\cite{nr_hosa}	&\textbf{0.9990}	&0.4705	&0.5925	&0.8574	&0.4494	&0.3248	&0.6412	&0.2993	&0.6393	&0.7399	&0.5851	&0.6275\\
& NIQE\cite{nr_niqe}	&0.9073	&0.3132	&0.6271	&0.8126	&0.3458	&0.2212	&0.6523	&0.5451	&0.7738	&0.5713	&0.5181	&0.5642\\
& MEON\cite{nr_meon}	&0.9409	&0.3750	&0.7248	&0.9215	&0.4101	&0.2497	&0.4861	&0.2980	&0.1917	&0.5466	&0.4969	&0.5570\\
& QAC\cite{nr_qac}	&0.8683	&0.3722	&0.4900	&0.7686	&0.3196	&0.1944	&0.3239	&0.2272	&0.3579	&0.5524	&0.4292	&0.5338\\
\hline \multirow{3}{*}{$\begin{array}{c}\text {Empirical} \\
\text{Fusion}\end{array}$} & CISI\cite{fusion_cisi} & 0.9680 & 0.8150 & 0.9425 & 0.9270 & 0.8231 & 0.8063 & \textbf{0.9135} & 0.6920 & 0.8740 & 0.8612 & 0.8634 & 0.8831\\
& HFSIMc\cite{fusion_hfsim} & 0.9610 & 0.8228 & 0.9423 & 0.9205 & 0.7315 & 0.7982 & 0.8202 & 0.5075 & 0.8624 & 0.8453 & 0.8345 & 0.8550\\
& CM3\cite{fusion_cm} & 0.9207 & 0.7136 & 0.8073 & \textbf{0.9450} & 0.6452 & 0.7659 & 0.7114 & 0.5055 & \textbf{0.9206} & 0.7733 & 0.7575 & 0.6893\\
\hline \multirow{2}{*}{$\begin{array}{c}\text {Rank} \\
\text {Fusion}\end{array}$}
& RRF\cite{fusion_rrf, fusion_bliss} & \textbf{0.9736}	&0.6571	&0.8421	&0.9278	&0.7873	&0.7645	&0.4941	&0.3722	&0.7022	&0.8678	&0.7133	&0.7608\\
& RRFW\cite{fusion_weighted_rrf1} & 0.9648	&0.7864	&0.9245	&0.9080	&0.8055	&0.8296	&0.6153	&0.4240	&0.7352	&0.8606	&0.7874	&0.7632\\
\hline \multirow{3}{*}{\textbf{Proposed}} & \textbf{SF-ms} & 0.9659	&\textbf{0.8678}	&0.9467	&\textbf{0.9384}	&\textbf{0.8473}	&\textbf{0.8432}	&0.9027	&0.7963	&\textbf{0.8895}	&0.8553	&\textbf{0.8869}	&\textbf{0.9002}\\
& \textbf{RF-ms} & 0.9714	&\textbf{0.8684}	&\textbf{0.9619}	&\textbf{0.9402}	&\textbf{0.8443}	&\textbf{0.8584}	&0.9059	&0.7484	&0.8779	&\textbf{0.8732}	&\textbf{0.8897}	&\textbf{0.9036}\\
& \textbf{SF-m} & \textbf{0.9723}	&0.8376	&0.9506	&0.9331	&0.8379	&0.8065	&\textbf{0.9072}	&0.8150	&0.8799	&0.8624	&\textbf{0.8763}	&\textbf{0.8946}\\
\Xhline{2\arrayrulewidth}
\end{tabular}
\end{table*}

\begin{table}[t]
\scriptsize
\caption{Evaluation results when ``bad" IQA metrics are added to the list of models to be fused.}
\label{table:random&small}
\centering
\begin{tabular}{|c|c|cc|}
\Xhline{2\arrayrulewidth}
Type & Model & avg SRCC & avg PLCC \\
\hline \multirow{2}{*}{Rank Fusion}  & RRF\cite{fusion_rrf, fusion_bliss} & 0.6096	& 0.6467\\
& RRFW\cite{fusion_weighted_rrf1} & 0.7171	& 0.7353 \\
\hline \multirow{3}{*}{\textbf{Proposed}}   & \textbf{SF-ms} & \textbf{0.8859}	&\textbf{0.9003} \\
& \textbf{RF-ms} & \textbf{0.8842}	&\textbf{0.8987} \\
& \textbf{SF-m} & \textbf{0.8745}	&\textbf{0.8928} \\
\Xhline{2\arrayrulewidth}
\end{tabular}
\vspace{-0.5cm}
\end{table}

\vspace{-0.1cm}
\subsection{Amortized Inference}
In the original MAP formulation in Eq. \ref{eq:map}, the estimation of $\mathbf{z}$ is carried out iteratively on a dataset. This poses two challenges: it may not yield an accurate estimation when the scores of a singular image are available, and it tends to be time-consuming. To address these limitations, we introduce amortized inference, which simplifies the inference procedure without stressing the training. Instead of optimizing $\mathbf{z}$ directly, an encoder $e: \mathbb{R}^M \to \mathbb{R}$ is introduced to fuse the scores of a given image. The encoder can be parameterized as a Fully Connected Network (FCN) with weights $\theta_e$. During inference, the framework no longer needs a large dataset to do optimization. Instead, it requires only the $M$ scores of the testing image: $z_i = e(\mathbf{x_i};\theta_e)$. By substituting $z_i$ with $e(\mathbf{x_i};\theta_e)$ in Eq. \ref{eq:map_factorized}, the encoder is jointly optimized with other parameters. An overview of the framework is shown in Fig. \ref{fig:framework}. 


\vspace{-0.1cm}
\subsection{Rank Fusion}
Rank fusion offers a non-linear transformation of diverse score distributions to a consistent uniform distribution. Such a transformation helps to stabilize the fusion process. In this subsection, we show that rank fusion can be seamlessly integrated into our general framework. Rather than optimizing in the discrete space which is computationally intensive, we opt for a more effective solution by normalizing the discrete ranks and conducting our optimization in the continuous space. We compute the normalized rank $r_i^j \in \mathbb{R}$ corresponding to the original score $x_i^j$ by finding the index $R_i^j \in \mathbb{N}$ of $x_i^j$ in the ascendingly ranked $\mathbf{x^j} = \{x_i^j | i=1, 2, ..., N\}$. Subsequently, $r_i^j$ is derived by $r_i^j = R_i^j / N$. By replacing $x_i^j$ with $r_i^j$ in Eq. \ref{eq:map_factorized}, we formulate rank fusion as a special instance of the general framework. 


\section{Experiments}

\subsection{Implementation and Experimental Details}
To rigorously assess the fusion performance of the proposed method, we evaluate it on ten diverse subjectively rated IQA datasets. Six of them, LIVE R2 \cite{db_liveR2}, TID2013 \cite{db_tid2013}, CSIQ \cite{fr_mad_db_csiq}, VCL@FER \cite{db_vclfer}, CIDIQ50 and CIDIQ100 \cite{db_cidiq}, feature single distortion. Four others, MDID \cite{db_mdid}, MDID2013 \cite{md_sisblim_db_mdid2013}, LIVE MD \cite{db_livemd} and MDIVL \cite{db_mdivl} comprise multiply distorted images. To demonstrate the generalizability of the proposed method, we include 12 FR-IQA and 4 NR-IQA methods of diverse design philosophies \cite{duanmu2021biqa} and varying correlation w.r.t. MOS (see Table \ref{table:all}). The chosen metrics include both traditional methods and deep learning based methods. The metrics are VSI \cite{fr_vsi}, FSIMc \cite{fr_fsimc_ref_yiq}, IWSSIM \cite{fr_iwssim}, DSS \cite{fr_dss}, MCSD \cite{fr_mcsd}, CID MS \cite{fr_cid}, GMSD \cite{fr_gmsd}, FSIM \cite{fr_fsim}, SFF \cite{fr_sff}, QASD \cite{fr_qasd}, VIF \cite{fr_vif}, VIF DWT \cite{fr_dwtvif}, HOSA \cite{nr_hosa}, NIQE \cite{nr_niqe}, MEON \cite{nr_meon}, and QAC \cite{nr_qac}. The goal of this paper is to develop an IQA score fusion model, rather than a new IQA model. So we do no include individual IQA metrics outside of the fusion list in comparison. Nonetheless, our framework remains extendable, allowing for seamless integration of state-of-the-art metrics to enhance fusion results. Except for empirical fusion methods that use a predetermined set of IQA metrics, we use the same set of 16 metrics for all other fusion methods in all of our experiments. RRF \cite{fusion_rrf, fusion_bliss} and RRFW \cite{fusion_weighted_rrf1, fusion_weighted_rrf2} are the only two unsupervised learning-based IQA fusion models in the literature. We refer RRF as ``unsupervised learning-based" where the training set is used to calculate the ranked index of the testing image. For a fair comparison, unsupervised training-based methods are retrained on the large KADID10K \cite{db_kadid10k} dataset and tested on the ten testing sets mentioned above. Supervised learning-based fusion methods are not included in the comparison since the main focus of the paper is unsupervised fusion. To evaluate the performance, Spearman's Rank Correlation Coefficient (SRCC) and Pearson Linear Correlation Coefficient (PLCC) over the ten testing sets are provided in Table \ref{table:all}.

Our framework is versatile, allowing for diverse implementations of the encoder $e$, decoder $f$ and uncertainty estimation module $g$. The chosen implementations are used to demonstrate the general framework and are not necessarily the optimal configurations. For simplicity, we implement the encoder $e$ using a six-layer Fully Connected Network (FCN). The number of input and output channels of each layer is set to the number of models to be fused. LeakyReLU is added to each layer except for the last one. Finally, for each testing image, the FCN is used to predict a set of weights adaptive to the content to linearly combine the scores. This makes the encoder explainable since we can inspect the predicted weights. Due to its lightweight nature, the inference speed is generally very fast. Since the goal of most IQA metrics is to approximate human perception of quality which is measured by MOS or DMOS (Difference of MOS), it's natural to assume $f^j$ is generally monotonic w.r.t. the predicted MOS $z_i$. We implement $f^j$ as an exponential function in the form: $f^j(z_i^j) = - e^{a^j (z_i^j - b^j)} + c^j$. The uncertainty estimation function $g^j$ is chosen to be quadratic in the form: $g^j(z_i^j) = a^j{z_i^j}^2 + b^j z_i^j + c^j$. We train the parameters in $f$ and $g$ with other modules end-to-end using the Adam optimizer with a learning rate of $0.002$. We stop the training when the loss reaches a plateau.

\vspace{-0.2cm}
\subsection{Evaluation Results}

We introduce three variants of the proposed model: SF-ms, RF-ms, and SF-m. ``SF" stands for Score Fusion while ``RF" denotes Rank Fusion. The suffixes further detail the uncertainty estimation levels: ``-m" is Model-level while ``-s" stands for Score-level. ``-ms" means both levels of estimations are included. Table \ref{table:all} provides a comprehensive evaluation of the proposed model, five other fusion methods, and 16 individual IQA models used in the fusion across ten diverse testing datasets. All three versions of the proposed model outperform other fusion methods, as well as individual IQA metrics, in terms of average SRCC and PLCC. Owing to the fine-grained score-level uncertainty estimation, SF-ms and RF-ms rank among the top three in five and six out of ten individual testing sets, respectively. To further demonstrate this, an ablation study is conducted where the score-level uncertainty estimation is removed, resulting in the SF-m model. As anticipated, SF-m is inferior to SF-ms, although it surpasses other fusion methods on average, thanks to the accurate estimation of $f$ and model-level uncertainty. The predicted $g$ are also non-increasing for most IQA models, which aligns with our observation as shown in Fig. \ref{fig:cond_hist}. The predicted overall scale parameter $\tilde{\omega^j} \in \mathbb{R}^M$ of each IQA model is also negatively correlated to the SRCC and PLCC of individual models. Since the proposed framework can easily be extended to rank fusion, we include RF-ms in comparison. This also provides a fair testing environment for other fusion-based methods, such as RRF and RRFW, since the scores are converted to ranks in the same way. As shown in the table, RF-ms outperforms other rank fusion methods, demonstrating the superiority of the proposed framework. 


To further demonstrate the importance of both model-level and score-level uncertainty estimation, we carried out an additional experiment where two ``bad-performing" IQA metrics are included in the fusing list. Without any prior knowledge of these metrics, many unsupervised learning-based methods may suffer since distinguishing ``bad" metrics becomes challenging without MOS or precise uncertainty estimation. We implement the two ``bad" metrics using random number generators producing uniformly sampled numbers from 0 to 1, which is the most common range for FR-IQA models. The results are shown in Table \ref{table:random&small}. The performances of RRF and RRFW decline significantly in this case, while the performance of the proposed method remains competitive. A closer examination of the uncertainty estimation module reveals that the estimated overall scale parameters $\tilde{\omega^j}$ of the two ``bad" metrics are significantly higher than others. As a result, the encoder $e$ assigns very little weight to these two metrics.

\section{Conclusion}
We propose a general framework for unsupervised IQA score fusion using MAP estimation. To the best of our knowledge, it is the first unsupervised learning-based score fusion approach for IQA. The proposed model conducts fine-grained uncertainty estimation at the score level to increase the accuracy. Rank fusion can be seamlessly integrated into the framework. The proposed model also exhibits interesting robustness at automatically rejecting ``bad" IQA models in the fusion process. 


\bibliographystyle{IEEEbib_abbrev}
{\footnotesize
\bibliography{IEEEabrv, refs}}

\end{document}